\documentclass[runningheads]{llncs}
\usepackage{graphicx}
\graphicspath{{imgs/}}
\usepackage{subfig}
\begin{document}
\title{Analyzing EEG Data with Machine and Deep Learning: A Benchmark}
\titlerunning{Analyzing EEG Data with Machine and Deep Learning: A Benchmark}
%

\author{
 Danilo Avola\inst{1} \and
 Marco Cascio\inst{1} \and
 Luigi Cinque\inst{1} \and
 Alessio Fagioli\inst{1} \and \\
 Gian Luca Foresti\inst{2} \and
 Marco Raoul Marini\inst{1} \and
 Daniele Pannone\inst{1}
}
%
\authorrunning{D. Avola et al.}
%

 \institute{
     Sapienza University of Rome\\
    \email{\{avola,cascio,cinque,fagioli,marini,pannone\}@di.uniroma1.it}  \and
     Università degli Studi di Udine\\
    \email{gianluca.foresti@uniud.it}
 }
\maketitle              
\begin{abstract}
Nowadays, machine and deep learning techniques are widely used in different areas, ranging from economics to biology. In general, these techniques can be used in two ways: trying to adapt well-known models and architectures to the available data, or designing custom architectures. In both cases, to speed up the research process, it is useful to know which type of models work best for a specific problem and/or data type. By focusing on EEG signal analysis, and for the first time in literature, in this paper a benchmark of machine and deep learning for EEG signal classification is proposed. For our experiments we used the four most widespread models, i.e., multilayer perceptron, convolutional neural network, long short-term memory, and gated recurrent unit, highlighting which one can be a good starting point for developing EEG classification models.

\end{abstract}
\section{Introduction}
\label{sec:introduction}
In the last years, machine learning (ML) and deep learning (DL) techniques have been used to overcome the limitation of classical image processing approaches in several fields \cite{avola2018features,dong2021survey}. The prominent results allowed researchers and industries to get outstanding outcomes in different tasks, such as environmental monitoring \cite{avola2016robot,patricio2018environmental,avola2019slam}, medical and rehabilitation \cite{petracca2015rehabilitation,avola2021thyroid,avola2009encephalic}, deception detection \cite{avola2019deception}, and more \cite{avola2017adaptive}. In addition, ML and DL have also found application for cross-domain problems, such as trading, biology, and neuroscience. Despite these very different contexts, an ML/DL practitioner commonly follows a well-defined pipeline. The latter usually consists in: 1) choosing a model (or designing one from scratch), 2) setting its hyperparameters, 3) training the chosen model, and finally 4) evaluating it. Then, if the observed results are not convincing, the process is re-iterated from 2) and, in the worst case, even from 1). Hence, to avoid changing or redesigning the model, in order to speed up the process, it is useful to know a priori the class of models that works the best for a certain problem or typology of data. 

For this reason, in this paper a benchmark on ML and DL models for EEG classification is proposed. The motivation of providing a benchmark for such a task is twofold. Firstly, EEG signals classification is increasingly used for different applications, such as security and smart prosthesis \cite{pham2013eeg,ruhunage2017eeg}. Secondly, to the best of our knowledge, there are no benchmarks available that highlight which class of models works best with EEG data. The comparisons have been performed by using four well-known ML and DL techniques, namely, the multilayer perceptron (MLP) \cite{557673}, the convolutional neural network (CNN) \cite{yildirim2020deep}, the long short-term memory (LSTM) \cite{wang2018lstm}, and the gated recurrent unit (GRU) \cite{chen2019hierarchical}. These approaches have been tested on the EEG motor movement/imagery dataset \cite{schalk2004bci}, which contains EEG data of 109 volunteers performing different tasks.

The remainder of this paper is organized as follows. Section \ref{sec:related} reports the current state-of-the-art in EEG signal classification techniques. Section \ref{sec:method} provides a background on EEG signals and on the methods used for this benchmark. Section \ref{sec:experiments} presents the performed experiments on the chosen dataset, highlighting the best typology of models to perform EEG signals classification. Finally, Section \ref{sec:conclusions} concludes the paper. 
\section{Related Work}
\label{sec:related}

In the last decades, several methods have been developed to analyze EEG signals and try to address different problems. 
Effective pipelines 
generally perform data pre-processing procedures, to reduce noise and to clean signals, a feature extraction step, to obtain a meaningful input representation, and an inference phase to correctly associate signals to the addressed task.


Concerning data pre-processing procedures, several techniques can be employed to remove artifacts from signals, such as filtering methods or wavelet transforms \cite{jiang2019removal}. In \cite{wang2018lstm}, for instance, MI-BCI signals are improved via the fast Fourier transform (FFT) and stacked restricted Boltzmann machines (RBM). 
The refined signals can then be fed to a CNN to extract meaningful features, which are in turn analyzed and classified via an LSTM network. 
Differently, in \cite{rabcan2020review}, the authors explore various procedures, e.g., discrete wavelet or Fourier transforms, to polish the input signals. Subsequently, other techniques, i.e., PCA and fuzzification, are also applied as pre-processing to, respectively, reduce the input dimensionality and prepare the data for a fuzzy decision tree (FDT) classifier to detect epileptic seizures.
While these approaches can be beneficial when analyzing EEG signals, other works concentrate on raw data by defining specific architectures. In \cite{zhang2017multi}, for example, an autoencoder is designed to reduce signal noise by exploiting brain activity of multiple persons. The XGBoost algorithm is then applied to these improved signals for MI classification. 

Regarding the feature extraction step, it enables to capture diverse aspects from BCI signals and can be performed in many different ways. 
In \cite{wang2018deep}, for example, an autoregressive discrete variational autoencoder (dVAE) is implemented to retain dynamic signal information. 
A Gaussian mixture hidden Markov model (GM-HMM) is then used to check for open or closed eyes. 
The authors of \cite{li2017epileptic}, instead, detect epileptic seizures using an SVM by means of multiscale radial basis functions (MRBF), and a modified particle swarm optimization (MPSO), that can simultaneously improve the EEG signals time–frequency extracted features associated to seizures. Differently, the authors of \cite{luo2018exploring} employ a filter bank common spatial pattern (FB-CSP) in conjunction with a sliding window to extract spatial-frequency-sequential features that are used to classify MI by a gated recurrent unit (GRU) network.

In relation to the inference phase, it can be performed using various approaches such as machine or deep learning algorithms. In \cite{nguyen2017inferring}, for instance, a relevance vector machine (RVM) is used to predict speech from BCI signals represented through a Riemannian manifold covariance matrix. 
In \cite{zhou2016fuzzy}, instead, a support vector machine (SVM) is employed to classify gender discrimination, alcholism and epilepsy from k-means clustered inter/intra BCI channels dependencies, in the latter. 
As for DL methods, the authors of \cite{xu2019deep} address the MI task exploiting the transfer learning paradigm, and fine-tune a pre-trained architecture using short time Fourier transformed BCI signals. The latter transform is also used in \cite{tabar2016novel} to tackle the same MI task, although the authors introduce a custom model called CNN-SAE to automatically extract and combine time, frequency and location information from the input data. In \cite{lawhern2018eegnet}, finally, a compact depth-wise and separable CNN, which is one of the chosen models used in this work benchmark, is designed to be effective on separate tasks such as event related potential, feedback error-related negativity, movement related cortical potential, as well as sensory motor rhythm classifications.

To conclude this overview, although many works tend to develop a specific methodology, others try to present a common ground by comparing existing approaches on complex BCI EEG signals analysis tasks. 
The authors of \cite{garrett2003comparison}, for example, examine linear discriminant analysis (LDA), a MLP with a single hidden layer, and SVM performances on $5$ different mental tasks, i.e., rest, math, letter, rotate, and count. 
In \cite{gandhi2011comparative}, instead, the best combination to detect epileptic seizures is searched among several wavelet transforms, i.e., Haar, Daubechies, Coiflets, and Biorthogonal, and either a SVM or probabilistic neural network classifier (PNN).
Therefore, inspired by these works, and following their rationale, we present a benchmark with more advanced deep learning algorithms to address another task, i.e., the MI-BCI classification.
\section{Background and Method}
\label{sec:method}

To provide a clearer overview concerning the context of the benchmark, this section provides the necessary background needed to understand EEG signals. In addition, a summary of ML and DL approaches used in the benchmark is provided.

\subsection{Background on EEG Signals}
\label{subsec:background}

\begin{figure}[t]
	\centering
	\subfloat[]{\includegraphics[width=0.45\linewidth]{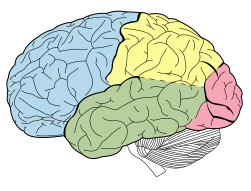}}
	\hspace{1pt}
	\subfloat[]{\includegraphics[width=0.45\linewidth]{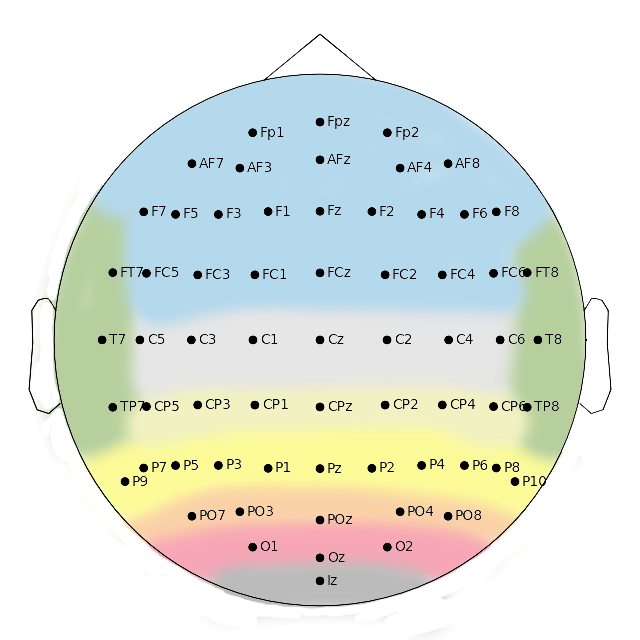}}
	\caption{The a) different areas of the human brain, and b) the corresponding 10-20 map system.}
	\label{1020}
\end{figure}

The first step regarding the analysis of EEG data is acquiring the latter during the execution of a specific task, i.e. a trial, from the subject involved in the experiment. This acquisition is usually performed by placing electrodes, which capture electrical signals emitted by human neurons, on specific positions of the subject scalp, as shown in Fig. \ref{1020}.
These signals have an extremely low voltage (between $-70mV$ and $+40mV$), where some peaks could determine if an action occurred. Indeed, the brain activity is determined by sequences of such peaks. The entire EEG spectrum is comprised in the range $[1Hz, 50Hz]$ and can be divided in several frequency bandwidths, i.e., the rhythms. Such rhythm, indicating different brain activities, are defined as: 

\begin{itemize}
    \item $\delta$ waves: representing deep sleep and pathological states, in the range of $[0.5Hz,3Hz]$;
    \item $\theta$ waves: indicating sleepiness and childhood memories, in the range of \\ $[3Hz,7Hz]$;
    \item $\alpha$ waves: representing adult relaxing state (rest status), in the range of $[8Hz,13Hz]$;
    \item $\beta$ waves: which identify the active state, involving calculus, focus, and muscular contraction, in the range of $[14Hz,30Hz]$;
    \item $\gamma$ waves: indicating tension or strong emotional states, over $30Hz$.
\end{itemize}

Usually, the acquired signal presents different types of noise. Examples of noise recurrent in EEG data is the muscular activity (such as eye blinks and movements), external sources of electricity, electrodes movements or even sounds. To sanitize the data, some well-known techniques, such as Individual Component Analysis (ICA) \cite{stone2002independent}, can be used. ICA decomposes the signals obtained from multiple electrodes during time in a sum of temporally independent (and fixed in terms of space) components. In particular, given an input matrix $X$ of EEG signals, where the rows represent the electrodes and the columns the time, the algorithm finds a matrix $W$ that decomposes and linearly smooths the data. The rows $U = WX$ represent the 
activation times of the ICA components, and have the form $components \times time$. The columns of the inverse matrix $W^{-1}$ provide the projection forces related to their respective components on each of the electrodes. In other words, these weights provide a topographical image of the scalp for each electrode. In this way, it is possible to locate brain areas involved in specific actions or feelings. 

\begin{figure}[t]
    \centering
    \includegraphics[width=0.5\linewidth]{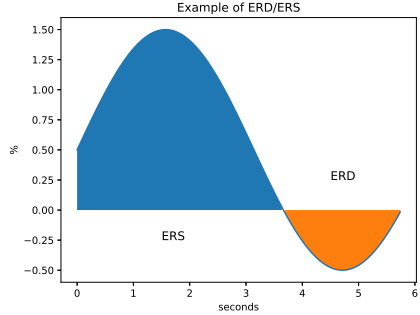}
    \caption{Example of ERS and ERD. With an ERS, we have an increment of the bandwidth power (filled in blue), while with an ERD we have a decrement of the bandwidth power (filled in orange).}
    \label{fig:erds_example}
\end{figure}

After data cleaning, it is easier to find the so-called \textit{event related potential} (ERP), which is the response to an internal or external event, namely, a sensory, a cognitive, or a motor event. In addition, these events can induce temporary modulations in certain frequency bandwidths, e.g. the increment or decrement of the bandwidth oscillation power. These modulations are called \textit{event related synchronization} (ERS) and \textit{event related desynchronization} (ERD). Synchronization and desynchronization terms refer to the fact that the temporary modulations are due to an increment or a decrement of the synchronization of a neuron population. An example of ERD and ERS is shown in Fig. \ref{fig:erds_example}.
Since the acquired EEG signal is a summary of the several brain activities, to get the signal regarding a specific stimulus, it is common practice to average the EEG signals obtained from the application of the same stimulus several times (e.g., 100 or more), allowing to remove disturbances not related to the stimulus itself.  

Remember that EEG signals are continuously recorded during the experiments. Hence, the resulting data is organized as a 2D matrix having the form $electrodes \times time$. To ease the analysis of EEG data in task-related experiments, usually the process that goes under the name of \textit{epoching} is performed. The latter consists in re-arranging the 2D data in a 3D data matrix, having the form $(electrodes,time,events)$.

\begin{figure}[t]
	\centering
	\includegraphics[width=1\linewidth]{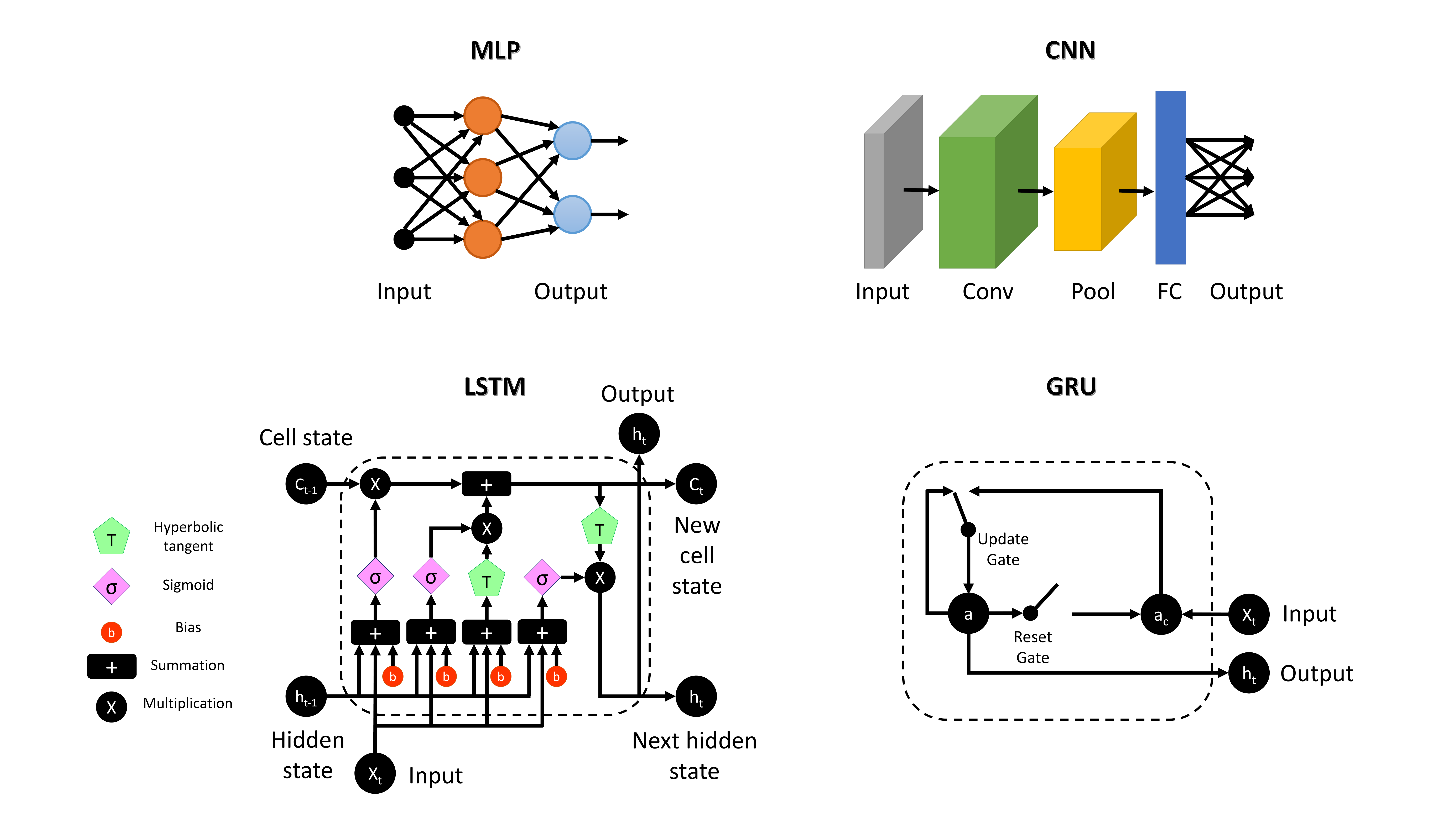}
	\caption{Simplified examples of the used architectures. From top left to bottom right: multilayer perceptron (MLP), convolutional neural network (CNN), long short-term memory (LSTM) and the gate recurrent units (GRU). The depicted architectures represent the base structure of each model, meaning that only the main elements are present.}
	\label{architectures}
\end{figure}
\subsection{Summary on used Machine and Deep Learning Approaches}
\label{subsec:method}
For the proposed benchmark, four well-known approaches were tested, namely, MLP, CNN, LSTM, and GRU. These approaches were chosen since they are the most representative of their category: MLP for classical ML, a standard CNN for DL approaches that are not designed to work with temporal data, and LSTM and GRU for DL approaches designed to work with temporal data. As it is possible to see, two models that can correctly handle temporal data were chosen. This is due to the fact that the final accuracy of those models strongly depends on the input data, therefore there is no better model between them. In Fig. \ref{architectures}, examples of the chosen approaches are shown.

The first approach we have chosen is MLP, which belongs to the class of feedforward artificial neural networks, and is the most basic neural network approach that can be used in classification task. The simplest MLP consists of three layers, namely, an input layer, a hidden layer, and the output layer. In the proposed benchmark, the used MLP has 1 input layer, 3 hidden layers, and 1 output layer.

CNN, originally designed for image analysis, can be used for analyzing EEG without efforts. As the name suggests, CNN uses convolutions in order to extract features from the input data. Depending on the data type, CNN may use 1D, 2D, or even 3D convolutions. In our benchmark, we used 1D convolutions with kernels of size $(channels,1)$. This kernel size allowed analyzing contemporary all the channels at each time instant. The used CNN consists of 4 convolutional layers and 1 dense layer for classification.

LSTM, differently from MLP and CNN, has the purpose of handling temporal data. Belonging to the family of recurrent neural networks (RNN) \cite{rumelhart1985learning}, a single LSTM unit is composed of a cell, that contains values over time intervals, and several gates, namely, an input gate, an output gate, and a forget gate. The aim of these gates is to control the flow of the information into and out of the cell. This structure allows to overcome the vanishing gradient problem that affects the training of standard RNNs. The LSTM used in the proposed benchmark simply consists of 2 stacked LSTM units.

The last approach we are going to summarize is the GRU. As for LSTM, GRU belongs to RNNs family. In fact, GRU can be seen as an optimized version of the LSTM due to its different internal structure. In detail, GRU has gating units allowing to control the information flow, but it lacks the memory cell. The less complex structure makes GRU computationally more efficient with respect to LSTM. As for the latter, in our benchmark we used 2 stacked GRU units.

\section{Experiments}
\label{sec:experiments}
In this section, the results of the benchmark, together with the used dataset, are presented. Concerning the data, the EEG signals were handled with MNE-Python, while all the tested methods were implemented in Pytorch. The machine used for training the model consisted in an AMD Ryzen 7 CPU, 16GB DDR4 of RAM, and an NVidia RTX 2080 GPU.
\begin{figure}[t]
	\centering
	\subfloat[]{\includegraphics[width=0.45\linewidth]{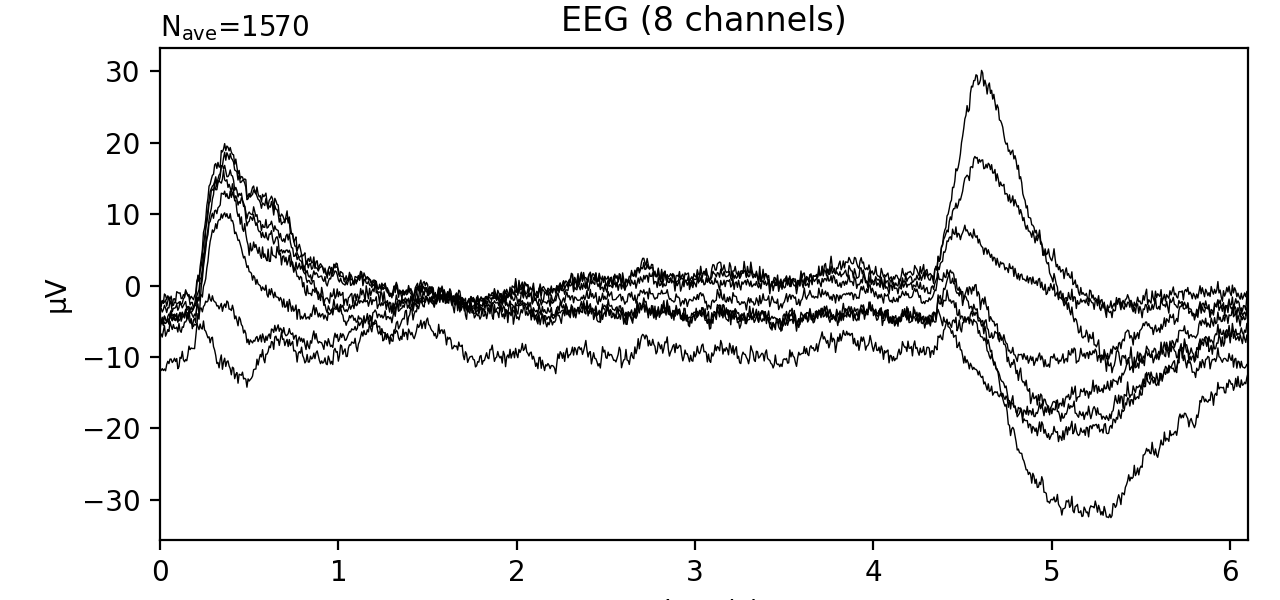}}
	\hspace{1pt}
	\subfloat[]{\includegraphics[width=0.45\linewidth]{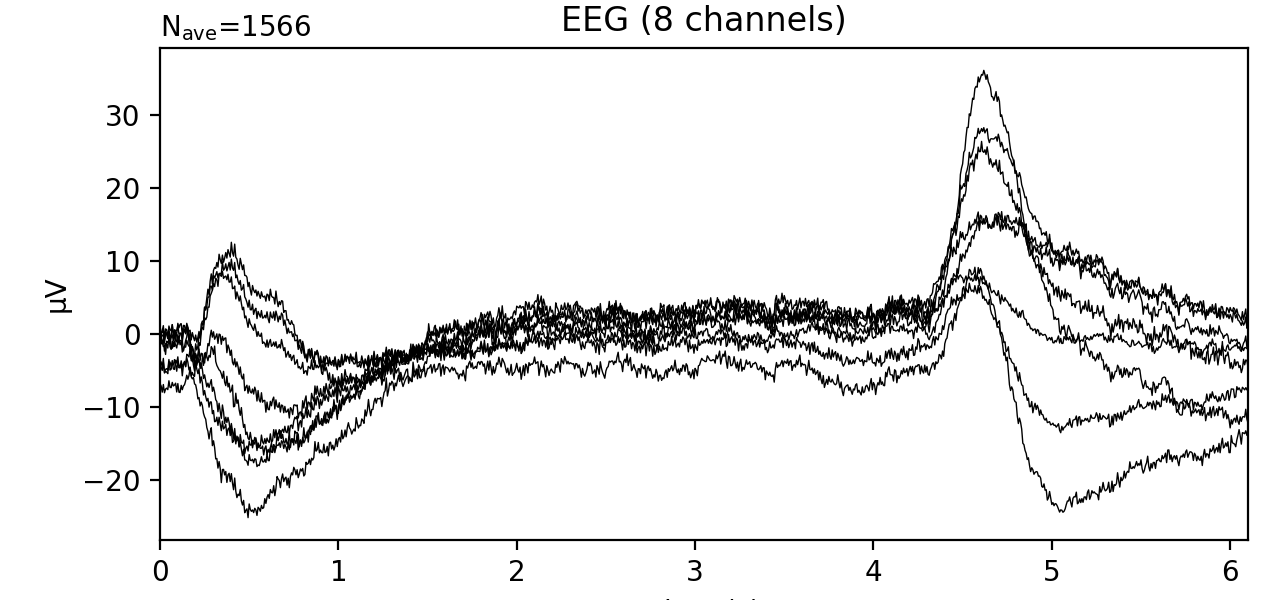}}
	\caption{Examples of signals corresponding to an event. In Figure a), the signal corresponding to the left hand is shown, while in b) the signal corresponding to the right hand is depicted.}
	\label{fig:evoked}
\end{figure}

\subsection{Dataset}
\label{subsec:dataset}
For the experiments, the EEG-BCI Motor Imagery Dataset \cite{schalk2004bci} has been used. It consists of more than $1500$ records in European Data Format (EDF) files from $109$ participants. The latter performed $14$ runs while wearing a BCI2000 device. At each run, the subject could perform one of the following actions: open or close right or left hand, imagine opening or closing right or left hand, open or close both hands or feet, imagine opening or closing both hands or feet. The EDF files contain $64$ EEG signals, with a frequency of 160 samples per second. In the experiments, we have chosen arbitrarily to classify signals related to right or left hand rather than distinguish signals of hands from feet. Before training the several models, the data has been cleaned from noise, and the data related to subjects having id $(43, 88, 89, 92, 100, 104)$ was removed. This is due to the fact that in some EEG signals the sampling frequency was different, or there was too much noise in the data. The final dimension of the dataset consisted in 3136 events to be classified. These events are then splitted as follows: 1980 events for training, 216 events for validation, and 940 for testing. Each event is fed as input in the form of a 2D matrix having the form $[channels, samples]$.

\subsection{Results}
\label{subsec:results}
In Fig. \ref{fig:evoked}, it is possible to see an example of signals related to the left (Fig. \ref{fig:evoked}a) and right (Fig. \ref{fig:evoked}b) hands. As it is possible to see, there are interesting patterns within the acquired signal. In detail, it is possible to observe that we have ERS in the intervals $[0,1]$ and $[4.5,5.4]$. Since these patterns are present in all the considered data, we choose to perform the experiments on different time intervals, which are the entire event, having length of $[0,6]$, and the two ERS specified above. In Table \ref{tab:results}, the results obtained with the different models are shown. As it is possible to see, the model that perform the best is the CNN, obtaining an accuracy of $90.4\%$ on the interval $[0,6]$, followed by MLP, with an accuracy of $85.2\%$. The worst results are obtained with the RNN models.
\begin{figure}[t]
    \centering
    \includegraphics[width=\linewidth]{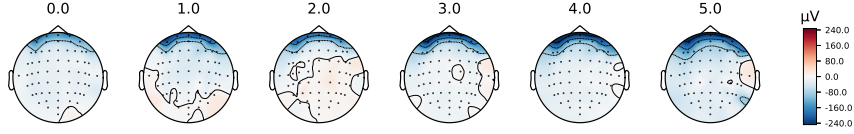}
    \caption{Activations of the electrodes during the runs. As it is possible to see, not every electrode provide relevant data. In this example, the electrodes on the top, i.e. the ones placed on the forehead, can be discarded since they acquire noisy data such as eye blinks.}
    \label{fig:evoked_plot}
\end{figure}
\begin{table*}[t]
    \centering
        \caption{Accuracy values obtained with the different models on the different intervals. In detail, we have that Interval 1 = $[0,1]$, Interval 2 = $[4.5,5.4]$, and Interval 3 = $[0,6]$.}
    \begin{tabular}{c c c c}
    \hline
    \hline
& \textbf{Interval 1} & \textbf{Interval 2} & \textbf{Interval 3} \\
\textbf{MLP} & 74.8\% & 84.6\% & 85.2\%\\
\textbf{CNN} & 78.8\% & 84.6\% & 90.4\% \\
\textbf{LSTM} & 65.4\% & 78.6\% & 57.9\%\\
\textbf{GRU} & 65.7\% & 85\% & 76.9\% \\
    \hline
    \hline
    \end{tabular}

    \label{tab:results}
\end{table*}
In detail, we have that GRU has obtained an accuracy of $76.9\%$, while the LSTM has obtained an accuracy of $57.9\%$, resulting as the worst model in our benchmark. Usually, RNNs performs better with respect to other methods when dealing with temporal or sequential data. This is amenable to the fact that the latter usually present some correlation among the several time instants. It is clear from the reported results that this aspect cannot be assumed in the case of brain signals. Instead, the several filters applied by the CNN seem to perform a better abstraction of the data, allowing to obtain the highest accuracy. This could be related to the fact that convolutions allow to better identify specific patterns, e.g. peaks, within the data.

During the experiments, we noticed that some channels provided more reliable data with respect to others. These channels are F7,F8, FT7, FT8, T9, TP7, TP8, and FC, as shown in the example in Fig. \ref{fig:evoked_plot}. Hence, we re-trained the chosen models by using only the above-mentioned channels, obtaining the results shown in Table \ref{tab:results_channels}.
\begin{table*}[t]
    \centering
        \caption{Accuracy values obtained by using only a subset of the channels. Also in this case, we have that Interval 1 = $[0,1]$, Interval 2 = $[4.5,5.4]$, and Interval 3 = $[0,6]$.}
    \begin{tabular}{c c c c}
    \hline
    \hline
& \textbf{Interval 1} & \textbf{Interval 2} & \textbf{Interval 3} \\
\textbf{MLP} & 77.3\% & 85.5\% & 85.7\%\\
\textbf{CNN} & 76.3\% & 85.2\% & 88.8\% \\
\textbf{LSTM} & 60.3\% & 82.2\% & 65.2\%\\
\textbf{GRU} & 67.1\% & 84\% & 78.7\% \\
    \hline
    \hline
    \end{tabular}

    \label{tab:results_channels}
\end{table*}
In this case, there is a general accuracy improvement among the several models and intervals. Despite this, we have that the CNN still results the best performing model with an accuracy score of $88.8\%$. Again, MLP follows with an accuracy of $85.7\%$. Finally, we have GRU and LSTM with an accuracy of $78.7\%$ and $65.2\%$, respectively.
\section{Conclusions}
\label{sec:conclusions}

In this paper, a benchmark of ML and DL models for EEG classification is presented. Four well-known models in ML and DL are compared, namely MLP, CNN, LSTM, and GRU. Extensive experiments have been performed on a dataset containing EEG data of 109 volunteers performing different tasks, highlighting that CNN is the best category of models for dealing with EEG data. 

\section*{Acknowledgments}
This work was supported by the MIUR under grant “Departments of Excellence 2018–2022” of the Sapienza University Computer Science Department and the ERC Starting Grant no. 802554 (SPECGEO).
%
%
%
\bibliographystyle{splncs04}
\bibliography{mybibliography}
\end{document}